\documentclass{article}
\usepackage[top=1in, bottom=1in, left=1in, right=1in]{geometry}
\usepackage[utf8]{inputenc} 
\usepackage[T1]{fontenc}    
\usepackage{hyperref}       
\usepackage{url}            
\usepackage{booktabs}       
\usepackage{amsfonts}       
\usepackage{nicefrac}       
\usepackage{microtype}      
\usepackage{graphicx}
\usepackage{todonotes}
\usepackage{amsmath}
\usepackage{bbm}
\usepackage{float} 
\usepackage{tikz}
\usepackage{placeins}
\usepackage{natbib}
\usepackage{setspace} 
\usetikzlibrary{patterns, shapes.geometric, positioning, arrows, shadows, matrix, fit, backgrounds}
\newcommand{\abbreviation}[2]{\textit{#2 (#1)}}
\DeclareMathOperator*{\argmax}{arg\,max} 
\DeclareMathOperator*{\PCA}{PCA} 
\DeclareMathOperator*{\OR}{OR} 
\DeclareMathOperator*{\FI}{FI} 
\DeclareMathOperator*{\II}{I} 
\DeclareMathOperator*{\RV}{RV} 
\DeclareMathOperator*{\RVa}{RVa} 
\newcommand{\ind}{\perp\!\!\!\!\perp}
\newtheorem{assumption}{Assumption}

\title{Management Decisions in Manufacturing using Causal Machine Learning -- To Rework, or not to Rework?}
\author{
  Philipp Schwarz\thanks{Equal Contribution.} \\
  University of Hamburg \\
  ams Osram \\
  \texttt{philipp.schwarz@ams-osram.com}
  \and  
  Oliver Schacht\footnotemark[1]\\
  University of Hamburg\\
  \texttt{oliver.schacht@uni-hamburg.de} \\
  \and
  Sven Klaassen\footnotemark[1]\\
  University of Hamburg \\
  Economic AI \\
  \and
  Daniel Grünbaum \\
  ams Osram \\
  \and
  Sebastian Imhof \\
  ams Osram \\
  \and
  Martin Spindler \\
  University of Hamburg\\
  Economic AI \\
}

\date{}

\begin{document}

\maketitle

\begin{abstract}
In this paper, we present a data-driven model for estimating optimal rework policies in manufacturing systems.
We consider a single production stage within a multistage, lot-based system that allows for optional rework steps.
While the rework decision depends on an intermediate state of the lot and system,
the final product inspection, and thus the assessment of the actual yield, is delayed until production is complete.
Repair steps are applied uniformly to the lot, potentially improving some of the individual items while degrading others.
The challenge is thus to balance potential yield improvement with the rework costs incurred.
Given the inherently causal nature of this decision problem,
we propose a causal model to estimate yield improvement.
We apply methods from causal machine learning,
in particular \abbreviation{DML}{double/debiased machine learning} techniques,
to estimate conditional treatment effects from data and derive policies for rework decisions.
We validate our decision model using real-world data from opto-electronic semiconductor manufacturing,
achieving a yield improvement of $2-3\%$ during the color-conversion process of white \abbreviation{LEDs}{light-emitting diodes}.
\end{abstract}
\singlespacing
\textbf{Keywords:} Causal Machine Learning, Heterogeneous Treatment Effects, Double Machine Learning, Policy Learning, Rework, Multistage Manufacturing, Production Optimization, Phosphor-converted White LEDs
\doublespacing
\section{Introduction}\label{Sec:intro}
Cost considerations and the increasing importance of sustainability 
require efficient manufacturing systems that produce high-quality items.
However, in complex value chains, such as semiconductor manufacturing,
imperfect processes can lead to products that fail to meet the required quality targets.
Several strategies have been proposed to avoid, reduce the number of, or cope with defective products
with the goal of minimizing costs.
While preventing defects is preferable to mitigating them, 
doing so usually assumes an in-depth knowledge of the processes involved (see \cite{Powell2022}).
In cases where this assumption does not hold, defect compensation techniques,
such as inline rework, seem more desirable than discard strategies.
We argue that in the case of rework, a well-informed economic decision should be made regarding
whether to repair a specific item. 
When dealing with an imperfect rework process in lot-based production systems,
the trade-off between the additional costs of and the savings from the repair step is even more apparent.
Indeed, depending on their state, some items in the production lot might benefit
from the repair while others deteriorate.
Thus, a decision model is needed that minimizes the yield loss incurred through the optional repair steps.

In the field of production planning and inventory control, rework has been intensively studied,
with a primary focus on improving the logistical decision-making associated with the rework process
(e.g., \cite{Wein1992,So1995,Alsawafy2022}).
Although logistical process properties such as lot size, production, and inspection cycles
affect the number of defectives (e.g., \cite{Lee1987}),
the actual impact of rework on the product has not been a major concern in this branch of research.
Except for \cite{Colledani2020}, the product state has usually been ignored,
leading to simplifications that do not allow for individual repair decisions.

Statistical quality control and improvement are more focused on defect avoidance.
The aim of research in this area is to increase quality by reducing process and product variability. 
This, in turn, is intended to decrease the number of defective items and the amount of rework \citep{Zantek2002}.
Recent quality models have jointly estimated the effect of process parameters and (intermediate) product measurements on process quality 
(e.g., \cite{Zantek2002,Senoner2022,Busch2022}).
However, policies derived from these models are usually of a global nature and thus do not allow for single-item decisions.

A direct approach to preventing and mitigating defects is taken in 
\abbreviation{ZDM}{zero defect manufacturing}.
Inline data-gathering techniques and analytics enable short feedforward cycles
that inform the next production step to compensate for variations introduced in the current step \citep{Powell2022}.
This approach allows for decision-making based on the current state of the item and system.
While the goal of ZDM is to avoid the production of scrap and thus the need to rework in the first place,
inline rework setups have been studied as defect mitigation measures \citep{Eger2018, Colledani2014a}.
This relates well to our work, in which defect detection and mitigation take place within a single
process stage. 

In all the abovementioned branches of research, the objective is to minimize the negative impact of imperfect production 
systems by taking actions upon some observed state.
Since any kind of control problem fundamentally involves identifying cause and effect relations,
this principle is relevant to our study as well, or as stated in \cite{Shewhart1926}:
\begin{quote}
  ``The reason for trying to find assignable causes is obvious -- it is only
  through the control of such factors that we are able to improve the
  product without changing the whole manufacturing process.''
\end{quote}
Despite this intrinsic causal reasoning, the predominant models in the rework literature do not emphasize it.
Modern correlation-based predictors excel at finding patterns in data
but falter when the underlying data-generating mechanisms change \citep{DAmour2020}.
Because optimizing a decision policy implies an imagined change in mechanism, 
we argue that robust causal methods 
should be used. 
\footnote{In a broad sense, causal (machine learning) methods refer to (machine learning) methods that utilize knowledge about the causal structure of the problem \citep{Kaddour2022}.}
We show that rework decisions are prone to confounding,
which can lead to spurious associations and a subsequent underestimation of their effect.
To overcome these drawbacks, we develop a data-driven framework for optimizing rework policies
based on methods from causal machine learning.

In particular, we rely on the double/debiased machine learning (DML) framework \citep{chernozhukov2018} to estimate the causal effect of rework on individual production lots.
Based on the estimate, we derive a counterfactual assignment policy for the rework step using conditional average treatment effects \citep{semenova2021debiased} and policy trees \citep{athey2021}.
In our setting, decision-making and treatment effect estimation target
the direct costs related to the final production yield instead of intermediate proxies of quality.
As argued by \cite{FernandezLoria2022}, this match in the outcome of interest is a precondition
for deriving optimal policies based on effect estimates.
Importantly, we are able to assess the influence of potential unobserved characteristics using bounds for omitted variable bias \citep{chernozhukov2023long}.

We gather empirical evidence of the effectiveness of our method by applying it to a real-world setting at AMS-Osram,
a leading manufacturer in the opto-semiconductor industry.
We investigate large-scale historic data of the inline rework process during the phosphor conversion 
of white \abbreviation{LEDs}{light-emitting diodes}.

Our main contributions are (1) the development of a data-driven framework for
optimizing rework policies in lot-based manufacturing systems using methods from causal machine learning
and (2) the empirical application of this framework to a challenging problem in opto-semiconductor manufacturing.

\section{Related Literature}\label{Sec:literature}
In this section, we provide an overview of the three main bodies of research relevant to our work.
First, we place rework in the general context of production planning, inventory control,
and quality management, and we relate it to ZDM approaches.
Second, we offer a short overview of the literature on causal approaches in manufacturing.
Lastly, we provide a brief introduction to the literature on the color-conversion process
during the manufacturing of white LEDs.

\subsection{Rework}
In the literature there are three dominant strategies for dealing with imperfect items in production:
avoiding defects, repairing defects, and managing imperfect production.
Early accounts of production planning and inventory control
that acknowledge the creation of defective items are extensions of the classical EOQ/EPQ Model from \citet{Harris1913}, such as those by \citet{Rosenblatt1986, Lee1987} and \citet{Porteus1986}.
\citet{Rosenblatt1986} assume that the production process deteriorates over time, transitioning from an ``in-control'' to an ``out-of-control'' state in which non-conforming items are produced.
Maintenance at the end of a production cycle resets the machine at some cost. 
The challenge is to find a cost-optimal production cycle time and corresponding lot size.
While this has some influence on the number of defective items,
the overall goal is a logistical one: managing the consequences of imperfect processes.
Several variations of and extensions to this model have been proposed, some of which deal with rework.
\citet{Wein1992} formulates a Markov decision process for finding optimal lot sizes in a make-to-order scenario with rework.
The repair step is assumed to eventually succeed, such that the desired number of good items is delivered to the next stage while the rest are scrapped. 
\citet{So1995} develop a policy for switching between regular and rework operations at a bottleneck stage
based on the buffer levels of the preceding rework and regular job queues.
\citet{Alsawafy2022} focus on single-item production with repeated rework and an imperfect repair process,
aiming to find the optimal production time for a single item. 
If this time is depleted, the continuous rework is stopped
and the current item is scrapped in favor of starting the production of a new one.
\citet{Colledani2020} propose a discrete-state Markov chain that models transitions in item 
and system states in a multi-stage production system with rework.
While the rework decision can be based on the item state, it is not the main objective of optimization.
Similar to the other research described above, the focus is on the logistics related to the rework process
rather than the actual rework decision.
Details on the occurrence of defectives and their repair are vastly simplified in these models.
Except for \cite{Colledani2020}, the state of an individual production lot is not considered.

An indirect but more process-centric strategy for avoiding defects is used in statistical quality control and improvement.
Although the primary goal of this strategy is to increase output quality by decreasing process and product variability, it is assumed that doing so decreases the number of defective items as well (e.g., \cite{Zantek2002}).
As early as \citet{Shewhart1926}, the causal nature of this endeavor was understood,
with the aim being to find assignable causes that explain variations in output quality and subsequently to control them.
Since then, various data-driven models have been developed to
(1) predict quality variation given process and product parameters,
and (2) exploit the learned models to minimize quality variation.
\citet{Zantek2002} develop a linear model to estimate the variation in production induced by a single production stage
on the final product in a multi-stage manufacturing system.
The model is used to guide the management investment in process improvement.
\citet{Senoner2022} learn a metamodel to predict variation in yield in a multi-stage system given a
set of production parameters.
They use techniques from explainable AI to decompose the learned model
and identify actions on the production parameters that minimize yield variation.
\citet{Busch2022} similarly derive limits for production parameters to increase yield,
employing different machine learning models.
Although the models in this line of work are conditional on the product and system state,
the decision-making is usually at the system level and thus affects multiple lots at once.
For example, \citet{Senoner2022} achieve a decrease in yield variation through a change in machine routing,
which simultaneously affects several production lots.
Moreover, non-causal machine learning models are usually employed to estimate the effect of causal interventions on process parameters.
Remarkably, the model assumptions in \citet{Zantek2002} resemble those of linear \abbreviation{SCMs}{structural causal models} (see e.g., \cite{Peters2017}). 

More direct approaches to defect avoidance and mitigation can be found in ZDM literature.
ZDM is a production paradigm that aims to eliminate scrap completely through 
defect prediction and prevention measures \citep{Powell2022}.
Short feedforward cycles are used to issue corrective measures, adjusting the next production steps
to prevent defective items. 
The correction decisions are usually based on the current product and system state,
which is inferred through inline inspection techniques.
Although rework in general is not in accordance with the overall goal of ZDM
to prevent defects from the outset,
inline rework has been studied as a defect compensation measure (e.g., \cite{Colledani2014a, Eger2018}). 
It involves defect detection and mitigation measures in the same manufacturing stage
without the expense of a setup apart from the main line \citep{Colledani2014a}. 
In our setup as well, the downstream compensation of defects takes place in the same stage 
and involves adjusted processing parameters.
\cite{Eger2018} developed a system that combines different data sources capturing system and product states
and carries out inline rework as a defect compensation measure, among other ZDM techniques.
\cite{Colledani2014a} apply inline rework during the manufacturing of electrical drives,
as part of which magnetic field deviations are compensated through the realignment of parts of the rotor.
Most ZDM approaches are process- or product-centric, whereas human-centric ZDM systems have hardly been researched \citep{Powell2022}.
Although we do not regard our approach to rework as human-centric,
in the application part of our paper we aim to reverse-engineer and improve the
decision policy of operators to aid future rework decisions.
In fact, without the human-made variation in the data, improving the in-sample policy would not be possible.
While \cite{Powell2022} list \abbreviation{AI}{artificial intelligence} for automated decision-making as a key enabling technology in ZDM,
the emerging trend in AI towards causality is insufficiently reflected in this field.

\subsection{Causal Machine Learning in Manufacturing}
Since the popularization of consistent causal reasoning through \citet{rubin2005} and \citet{pearl1995},
both the potential outcomes framework and graph-based approaches have influenced various areas of research.
The recent combination of causality with machine learning aims to circumvent shortcomings in traditional ML methods, such as biased estimates due to confounding.
For a thorough overview, see \citet{Kaddour2022}.

Causal machine learning has been applied to a broad variety of settings,
ranging from physical (e.g., \cite{Schoelkopf2015}) to marketing science problems (e.g., \cite{Ellickson2023}).
In manufacturing, several survey articles have referenced causal methodology.
\cite{Diedrich2020} propose a concept for automated fault diagnosis and reconfiguration for hybrid cyber-physical production systems based on a causal dependency model.
\cite{Hua2022} develop a combined framework of deep convolutional learning and causal representation learning, which they apply to the prediction of tool wear under non-stationary working conditions.
More generally, \cite{Hnermund2021} provide an overview of opportunities for using causality in business decision-making.
Lastly, \cite{Ho2017} give an overview of multiple causal methods and propose their application in operations management.
However, to the best of our knowledge, there is no application of heterogeneous treatment effect estimation and policy learning in a manufacturing context.
Furthermore, there is no work on the application of causal machine learning to rework decision-making
in manufacturing beyond preliminary studies conducted by \cite{Schacht2023, Gruenbaum2023}.

\subsection{Color Conversion}
For the interested reader, we provide a short introduction to the literature 
regarding the physics of color-converted LEDs.
Although the development of LEDs is an interesting and active area of research
in its own right (e.g., \cite{Wankerl2022}), it is relevant here only for the application part of our work.
\citet{Cho2017} provide a thorough account of the history of polychromatic LEDs.
The work of \citet{Schubert2014} can be regarded as a standard reference for general LEDs.
According to \citet{Cho2017}, the prerequisites for polychromatic white LEDs were established in 
the mid-1990s with the development of LEDs covering the entire monochromatic spectrum.
Since then, different approaches to polychromatic white LEDs have been developed.
One commercially successful method involves dispensing a phosphor conversion layer on monochromatic blue LEDs.
Although this process step is crucial for the optical performance of the LEDs \citep{Wang2014,Lo2014},
it is difficult to control in large-scale manufacturing due to various interrelated physical mechanisms.
For example, layer thickness influences perceived color,
and affects temperature and thermodynamic stress in the LED \citep{Tan2018}.
Furthermore, the sedimentation of phosphor particles in the dispensed slurry influences 
luminous efficiency and color homogeneity (e.g., \cite{Wang2014}).
Various dispensing setups have been studied in physics-orientated literature (e.g., \cite{Lo2014,Huang2010,You2010}). 
In contrast, we adopt a high-level, data-driven perspective that is more suitable for manufacturing.
Put differently, we assume that the setup of the physical process and the corresponding parameters are fixed
and reasonably refined.

\section{Model}
\FloatBarrier
The problem setup discussed in the following is derived from an application in opto-semiconductor production.
For clarity of presentation and to highlight the broad applicability of our framework,
we first describe the abstract manufacturing setting and
the applied data-driven methods before presenting the empirical results.

\subsection{Manufacturing Setting}
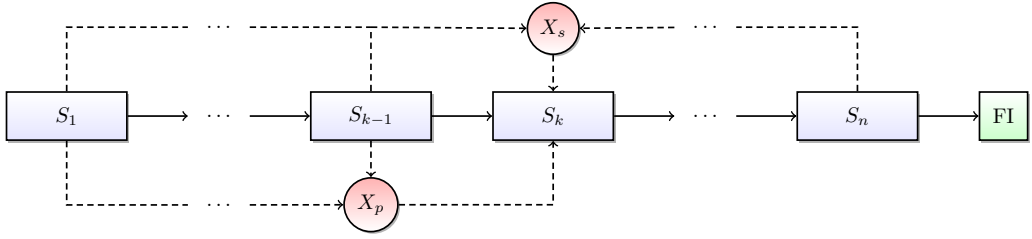
\begin{figure}
\centering
  \scalebox{0.8}{\begin{tikzpicture}
  \begin{scope}[
    thick,
    every node/.style={minimum width=0.8cm, minimum height=0.8cm},
    stage/.style={rectangle,draw,top color=white,bottom color=blue!10,general shadow={fill=gray!60,shadow xshift=1pt,shadow yshift=-1pt},minimum width=2cm},
    mea/.style={circle,draw,top color=red!30,bottom color=white!10,general shadow={fill=gray!60,shadow xshift=1pt,shadow yshift=-1pt}},
    meaout/.style={rectangle,draw,top color=white,bottom color=green!20,general shadow={fill=gray!60,shadow xshift=1pt,shadow yshift=-1pt}},
    hidden/.style={rectangle,fill=white,text opacity=1,fill opacity=0, minimum width=1.0cm},
    infoflow/.style={densely dashed},
    productflow/.style={},
  ]
    \node[stage] (S1) { \(S_1\) };
    \node[hidden,right=of S1] (DOTS1) { \(\ldots\) };
    \node[stage,right=of DOTS1] (SKm1) { \(S_{k-1}\) };
    \node[stage,right=of SKm1] (SK) { \(S_k\) };
    \node[hidden,right=of SK] (DOTS2) { \(\ldots\) };
    \node[stage,right=of DOTS2] (SN) { \(S_n\) };

    \draw [productflow,->] (S1) edge (DOTS1);
    \draw [productflow,->] (DOTS1) edge (SKm1);
    \draw [productflow,->] (SKm1) edge (SK);
    \draw [productflow,->] (SK) edge (DOTS2);
    \draw [productflow,->] (DOTS2) edge (SN);

    \node[mea,below=0.6cm of SKm1] (XP) { \(X_p\) };
    \node[hidden,below=0.65cm of DOTS1] (DOTSXP) { \(\ldots\) };
    \path[infoflow,draw,-] (S1) |- (DOTSXP);
    \path[infoflow,draw,->] (DOTSXP) edge (XP);
    \path[infoflow,draw,->] (SKm1) edge (XP);
    \path[infoflow,draw,->] (XP) -| (SK);

    \node[mea,above=0.6cm of SK] (XS) { \(X_s\) };
    \node[hidden,above=0.65cm of SKm1] (ARCXS) {};
    \node[hidden,above=0.65cm of DOTS1] (DOTSXS1) { \(\ldots\) };
    \path[infoflow,draw,-] (S1) |- (DOTSXS1);
    \path[infoflow,draw,-] (DOTSXS1) edge (ARCXS.center);
    \path[infoflow,draw,->] (ARCXS.center) -> (XS);
    \path[infoflow,draw,-] (SKm1) -- (ARCXS.center);
    \path[infoflow,draw,->] (XS) edge (SK);

    \node[hidden,above=0.65cm of DOTS2] (DOTSXS2) { \(\ldots\) };
    \path[infoflow,draw,-] (SN) |- (DOTSXS2);
    \path[infoflow,draw,->] (DOTSXS2) edge (XS);

    \node[meaout,right=of SN] (YIELD) { \(\FI\) };
    \path[productflow,draw,->] (SN) edge (YIELD);


  \end{scope}
\end{tikzpicture}}
  \caption{
    Exemplary linear setup with manufacturing stages \(S_1, \ldots, S_n\) and a final inspection stage \(\FI\).
    Decisions at \(S_k\) may depend on previously observed product state \(X_p\)
    and the overall system state \(X_s\).
  }
  \label{model:fig:prod_flow}
\end{figure}

We assume that the observed value chain consists of multiple manufacturing stages (Figure \ref{model:fig:prod_flow}).
Each stage \(S_1,\, \ldots, \, S_n\) is composed of several process steps, including inline inspection of manufactured products.
Furthermore, we assume that the production is lot-based,
meaning that multiple items are grouped together and receive the same treatment
while traversing the production stages.
Any decision or parameter adjustment at a specific process step affects the current lot as a whole,
and no item-level decisions can be made.
After a lot has passed the final production stage \(S_n\), a conformity check \((\FI)\) is performed,
which determines the fraction \(Y\) of acceptable items.

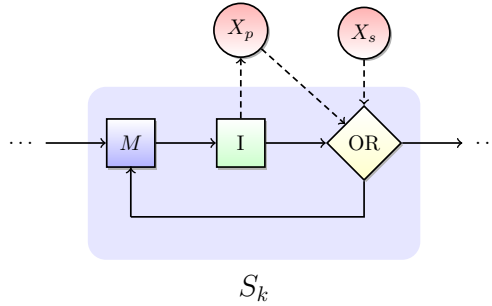
\begin{figure}
\centering
  \scalebox{0.8}{\begin{tikzpicture}
  \begin{scope}[
    thick,
    local bounding box=graph,
    every node/.style={minimum width=0.8cm, minimum height=0.8cm},
    pstep/.style={rectangle,draw,top color=white,bottom color=blue!30,general shadow={fill=gray!60,shadow xshift=1pt,shadow yshift=-1pt}},
    instep/.style={rectangle,draw,top color=white,bottom color=green!20,general shadow={fill=gray!60,shadow xshift=1pt,shadow yshift=-1pt}},
    decision/.style={diamond,draw,top color=white,bottom color=yellow!20,,general shadow={fill=gray!60,shadow xshift=1pt,shadow yshift=-1pt}},
    mea/.style={circle,draw,top color=red!30,bottom color=white!10,general shadow={fill=gray!60,shadow xshift=1pt,shadow yshift=-1pt}},
    hidden/.style={rectangle,fill=white,text opacity=1,fill opacity=0},
    infoflow/.style={densely dashed},
    productflow/.style={},
  ]
    \node[hidden] (DOTS1) {\(\ldots\)};
    \node[pstep,right=of DOTS1] (P1) { \(M\) };
    \node[instep,right=of P1] (P2) { \(\II\) };
    \node[decision,right=of P2] (DEC) { \(\OR\) };
    \node[hidden,right=of DEC] (DOTS2) {\(\ldots\)};
    \node[mea,above=of P2] (XP) {\(X_p\)};
    \node[mea,above=0.75cm of DEC] (XS) {\(X_s\)};

    \draw[productflow,->] (DOTS1) edge (P1);
    \draw[productflow,->] (P1) edge (P2);
    \draw[productflow,->] (P2) edge (DEC);
    \draw[productflow,->] (DEC) edge (DOTS2);

    \draw[infoflow,->] (P2) edge (XP);
    \draw[infoflow,->] (XP) edge (DEC.north west);
    \draw[infoflow,->] (XS) -- (DEC);

    \node[hidden,below=0.4cm of P1] (RWARC) { };
    \path[productflow,draw,-] (DEC.south)  |- (RWARC.center);
    \path[productflow,draw,->] (RWARC.center) |-  (P1.south);


  \end{scope}
  \begin{scope}[
    on background layer,
    thick,
    process/.style={fill=blue!10,rectangle,rounded corners=2ex},
  ]
  \node[process,label={[thick,shift={(0.0,-3.7)}]\Large\(S_k\)}, fit=(P1) (P2) (DEC) (RWARC), inner sep=3mm] (SK) {};
  \end{scope}
\end{tikzpicture}}
  \caption{
    Manufacturing \(M\), inline inspection \(\II\) and decision step \(\OR\)
    at stage \(S_k\).
  }
  \label{model:fig:stage_flow}
\end{figure}

For decision-making, we focus on an intermediate stage \(S_k\),
at which we have access to a set of values \(X_p\) and \(X_s\) derived
from the actual but unknown product state \(P\) and system state \(S\).
The observed product state \(X_p\) is measured through inline inspections performed at the 
current and previous stages. It captures measurements of individual items in the current lot,
whereas the observed system state \(X_s\) consists of features 
related to the state of the human operators (e.g., time until break)
and the production equipment (e.g., overall load, contamination, time until next maintenance).
Although both parts comprise the observed state \(X = (X_p, \, X_s)\),
\(X_s\) cannot be influenced through any decision-making at \(S_k\).
Of course, in reality, the observed state depends on the decision-maker.
For example, the human decision-maker might implicitly be aware of and affected by the system load.
This is in contrast to fully automated environments, where process steps are exactly repeated in strict timing.
\(S_k\) is supposed to admit the possibility of inline rework (see Figure \ref{model:fig:stage_flow}).
A process step \(M\) is followed by an inline inspection \(\II\),
which updates the observed product state \(X_p\).
Subsequently, a decision \(A \in \{0, 1\}\) is made whether to repeat \(M\) 
with adjusted process parameters
or to continue with the next production stage.
Parameter settings for the initial regular step and
the possible repeated rework steps are assumed to be fixed.

While each defective item in a lot implies a certain cost, so does the additional rework step.
Because the rework step is done inline, we assume that the setup costs for
switching back and forth between regular and rework operation modes
are negligible compared to the process costs for a single rework step.
Further we suppose that the cost of imperfect production is linear with respect to the number of defectives.
This allows us to express the total costs \(c\) for a rework step in units of yield percent.

\subsection{Causal Setting}\label{model:sec:causal_setting}
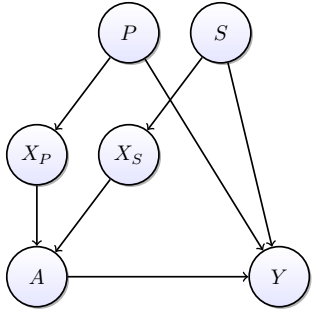
\begin{figure}
\centering
  \scalebox{0.8}{\begin{tikzpicture}
  \begin{scope}[
    thick,
    every node/.style={minimum width=1cm, minimum height=1cm},
    decision/.style={diamond,draw,top color=white,bottom color=yellow!20,,general shadow={fill=gray!60,shadow xshift=1pt,shadow yshift=-1pt}},
    block/.style={circle,draw,top color=white,bottom color=blue!10,general shadow={fill=gray!60,shadow xshift=1pt,shadow yshift=-1pt}},
    hidden/.style={rectangle,fill=white,text opacity=1,fill opacity=0},
  ]
    \node[block] (SYS_STATE) { $S$ };
    \node[block,left=0.5cm of SYS_STATE] (PROD_STATE) { $P$ };

    \node[block,below=of PROD_STATE] (OBS_SYS_STATE) { $X_S$ };
    \node[block,left=0.5cm of OBS_SYS_STATE] (OBS_PROD_STATE) { $X_P$ };

    \node[block,below=of OBS_PROD_STATE] (ACT) { $A$ };

    \node[block,right=3cm of ACT] (OUT) { $Y$ };

    \path [->] (SYS_STATE) edge (OBS_SYS_STATE);
    \path [->] (PROD_STATE) edge (OBS_PROD_STATE);

    \path [->] (OBS_SYS_STATE) edge (ACT);
    \path [->] (OBS_PROD_STATE) edge (ACT);

    \path [->] (PROD_STATE) edge (OUT);
    \path [->] (SYS_STATE) edge (OUT);
    \path [->] (ACT) edge (OUT);


  \end{scope}
\end{tikzpicture}}
  \caption{
    Graphical Causal Model:
    Yield \(Y\) and rework decision \(A\) are confounded by 
    lot state \(P\) and system sate \(S\).
  }
  \label{model:fig:dag}
\end{figure}

The above setup implies the causal relations shown in Figure \ref{model:fig:dag}.
Both the product state \(P\) and system state \(S\) before decision-making
affect the yield \(Y\) and the rework decision \(A\) --
the latter indirectly through the observed states \(X_p\) and \(X_s\).
The action \(A\), in turn, affects the yield \(Y\); otherwise, the rework step would be superfluous.
This makes \((P,\, S)\) a confounder of \(A\) and \(Y\),
meaning there is an open \textit{backdoor path} between \(A\) and \(Y\) \citep{pearl1995}.
It is likely that lots with severe defects
-- i.e., those beyond any repair -- tend to receive rework treatment more often
compared to those that have a chance of improvement but seem quite good to begin with.
Causal methods can exploit this knowledge to adjust for these kinds of confounding biases.
However, not all relevant parts of \((P,\,S)\) might be observed,
potentially disregarding some unobserved confounding influence \(U\).

\subsection{Problem Formulation}\label{sec:problem}
Our goal is to find a cost-optimal policy \(\pi^{\ast}\)
that decides based on the observed lot state \(X = (X_p,\, X_s)\)
whether to continue reworking \(\pi^{\ast}(X) = 1\) or
to proceed with the next production stage \(\pi^{\ast}(X) = 0\)
For any observed state \(X\), an optimal policy \(\pi^{\ast}\) should decide
for rework if the yield improvement \(\Delta Y\) due to this extra step 
outweighs the costs \(\Delta Y - c > 0\), but should continue with the next stage otherwise.
Importantly, neither the actual yield \(Y\) after production,
nor the yield improvement \(\Delta Y\) are known at the time of decision-making.
Calculating the latter involves estimating interventional quantities,
namely the yield of the lot when doing one additional rework step \(Y(A=1)\)
and the yield of the lot when skipping further rework steps \(Y(A=0)\).

Therefore, we assume that \(i=1, \ldots, N\) observations
\(W_i = (X_i, \, A_i, \, Y_i)\), each for an individual production lot, are given.
A data point \(i\) consists of
the observed system and product state \(X_i = (X_{i, p}, \, X_{i, s})\) before the rework decision \(A_i\),
the rework decision \(A_i\), and the yield \(Y_i\) after production is finished.

The causal interpretation of estimated quantities, such as \(\Delta Y\), requires the employed model to
account for the confounding influences of the product and system state via observed values $X$,
avoiding reliance on spurious associations \citep{Kaddour2022}.
Although the estimation of $\mathbb{E}[Y|A,X]$ (and with it \(\mathbb{E}[\Delta Y \,|\, X]\)) can be performed flexibly using naive machine learning methods,
these models are prone to regularization bias, necessitating the use of double machine learning.
Despite the adjustment for known confounding, these models (like every other model) are vulnerable to unobserved confounding.
Therefore, we also assess the sensitivity of the estimated models regarding unobserved confounders.


\section{Methods}
In the following, we describe the causal methodology we use for policy estimation.
We begin with an introduction to the potential outcomes framework,
followed by our approach to treatment effect estimation with double/debiased machine learning.
We also detail our treatment policy estimation based on orthogonal scores.
Lastly, we present our approach to sensitivity analysis for confounding factors.

\subsection{Causal Estimation Approach}\label{sec:causalinf}
In this section, we introduce our estimation procedure following the potential outcomes framework of \cite{rubin2005}.
We define $Y_i(a)$ as the potential outcome of a unit $i$ under the treatment decision $a\in \{0,\,1\}$.
$Y_i(a)$ is the potential yield of a production lot with rework \(a=1\) or without rework \(a=0\).
The individual effect of the treatment can be defined as
\begin{equation}
  Y_i(1)-Y_i(0)
\end{equation}
Unfortunately, due to the fundamental problem of causal inference \citep{Holland1986},
we can never observe the outcome for the same individual under treatment and not under treatment simultaneously.
We observe only treatment $A$, which is assigned by an (unobserved) mechanism $m: \mathcal{X}\rightarrow\mathcal{A}$ in the data. Here, $X_i\in \mathcal{X}$ are all characteristics of the production lot and the system state we observe, as explained in Section \ref{sec:problem}. \\

The fact that we cannot observe \textit{counterfactual} outcomes for an individual raises the question of how to estimate the effect of the treatment assignment. One common approach in the literature is to average over all individuals to identify the \abbreviation{ATE}{average treatment effect} \citep{Rubin1974} given by
\begin{equation}
\theta_0:=\mathbb{E}[Y(1)-Y(0)].
\end{equation}
Alternatively, to include the specific information (\textit{heterogeneity}) about individuals encoded in $\mathcal{X}$, the \abbreviation{CATE}{conditional average treatment effect}
\begin{equation}
\theta_0(z):=\mathbb{E}[Y(1)-Y(0)\mid Z=z]
\end{equation}
can be considered. Here, $Z$ can be a subset of $X$, representing a \abbreviation{GATE}{group average treatment effect} or additional covariates influencing the outcome but not the treatment decision.
In our setting, the average treatment effect is the effect of reworking every single production lot.
The conditional effect, however, is the effect conditional on some observable characteristics of the chip lot and system state. \\

The identification of $\theta_0$ and $\theta_0(z)$ strongly depends on the treatment assignment mechanism $m$. If we had a randomized assignment, known as a \textit{randomized controlled trial} or A/B-test, $\theta_0$ and $\theta_0(z)$ could be directly derived from the population means.
In a deterministic assignment setting (when $\mathbb{P}(A=1\mid X)\in\{0,1\}$), designs such as \textit{regression discontinuity} can be used to identify $\theta_{0}(C)$ at a cutoff level $C$. \\

We, however, base our effect identification on two main assumptions, as follows:
\begin{assumption}[Conditional Exchangeability]\label{ass:condex}
$Y(a)\ind A\mid X$
\end{assumption}
Assumption \ref{ass:condex} states that all dependencies between the potential outcomes and the assigned treatment are explained by $X$, so the confounding is completely explained by $X$. 
It is expressed graphically in Figure \ref{model:fig:dag} and enables us to identify the potential outcomes distributions with observable data.
\footnote{
  In particular, under Assumption \ref{ass:condex} \(P^*(Y(a) \leq y) = \int_{-\infty}^y \int_x f_{Y|A,X}(\hat{y} \,|\, a, \, x)f_X(x) dxd\hat{y} \)
holds with potential outcome distribution \(P^*\) and observable densities \(f_X\), \(f_{Y|A,X}\) \citep{Pearl2009}.
Naive estimation of the ATE using the observed cdf \(P(Y \leq y \,|\, A = a)\) instead of the 
corrected \(P^*(Y(a) \leq y)\) leads to the confounding bias described in Section \ref{model:sec:causal_setting}.
Note that when conditioning on all confounders, the naive estimator for the CATE coincides with the adjusted one
but becomes biased when leaving out some confounders
(e.g., conditioning on \(X_p\) but  not on \(X_s\), or disregarding unobserved confounders \(U\)).
}

\begin{assumption}[Overlap]\label{ass:olap}
$\exists \varepsilon>0: \varepsilon<\mathbb{P}(A=1\mid X)<1-\varepsilon \quad\forall X \in \mathcal{X}$
\end{assumption}
Assumption \ref{ass:olap} states that for the entire space of $\mathcal{X}$, units are assigned to treatment and control with a non-zero probability, enabling us to estimate the potential outcomes.\\

Based on these key assumptions, there are multiple well-known approaches to estimate the parameter $\theta_0$, such as inverse probability weighting, regression adjustment or matching.
In recent years, more sophisticated frameworks for estimating heterogeneous causal effects have been developed, such as targeted maximum likelihood estimation \citep{vanderLaanRubin+2006}, causal forests \citep{athey2021} and ML-based meta-learners (for an overview, see \citet{Knzel2019}).
In this work, we follow the recently developed semi-parametric double/debiased machine learning framework by \citet{chernozhukov2018}.\\

\subsection{Treatment Effect Estimation with Double/debiased Machine Learning}\label{sec:DML}
Unlike the most common use of modern machine learning algorithms for prediction or classification, the estimation of causal parameters such as the ATE with ML imposes additional challenges for the estimator.
As shown by \citet{chernozhukov2018}, biases from regularization or overfitting can be amplified,
leading to severe bias in the target estimate. \\
The work of \citet{chernozhukov2018} provides a framework that counteracts these issues,
enabling the use of algorithms such as random forests or boosting to obtain precise estimates of treatment effects combined with confidence intervals to access estimation uncertainty. \\

Under Assumptions \ref{ass:condex} and \ref{ass:olap}, the underlying causal structure can be represented by an \abbreviation{IRM}{interactive regression model}, which follows the SCM:
\begin{align*}
    Y &= g_0(A,X) + \xi,\quad \mathbb{E}[\xi\mid X, \, A] =0,\\
    A &= m_0(X) + \nu,\quad \mathbb{E}[\nu\mid X] = 0,
\end{align*}
where the conditional expectations 
\begin{align}
    g_0(A,X) &= \mathbb{E}[Y\mid X, A],\label{eq:g_0}\\
    m_0(X) &= \mathbb{E}[A\mid X] = \mathbb{P}(A=1\mid X)\label{eq:m_0}
\end{align}
are unknown and might be complex functions of $X$. In this structural equation model, the ATE 
\begin{equation*}\label{ATE}
    \theta_{0} = \mathbb{E}[g_0(1,X)-g_0(0,X)]
\end{equation*}
as well as the \abbreviation{ATT}{treatment effect of the treated}
\begin{equation*}\label{ATT}
    \theta_{ATT} = \mathbb{E}[g_0(1,X)-g_0(0,X)\mid A=1]
\end{equation*}
are identified. \\

In double machine learning, inference is based on a method of moments estimator
\begin{equation}
    E[\psi(W;\theta_0,\eta_0)] = 0,
\end{equation}
with $\psi(W;\theta_0,\eta_0)$ being a Neyman-orthogonal score that identifies the causal parameter $\theta_0$ given ML estimates of the parameters or functions $\eta_0$ (referred to as nuisance elements).
The Neyman-orthogonality property ensures the estimator is robust against small perturbations in the ML nuisance element estimates.
Furthermore, the use of $k$-fold crossfitting safeguards against overfitting affecting the estimation of the target parameter \(\theta_0\),
and the complexity of the estimator to be controlled, leading to appealing properties such as $\sqrt{n}$-consistency and approximate normality. \\

In case of an IRM, the score for an ATE estimator is given by the linear form
\begin{align}
	\psi(W_i;\theta,\eta) :=&\ \psi_a(W_i,\eta)\theta + \psi_b(W_i,\eta)
  \label{eq:psi}
\end{align}
with
\[
  \psi_b(W_i, \eta) :=  g(1,X_i) - g(0,X_i) \nonumber + \frac{A_i(Y_i - g(1,X_i))}{m(X_i)} - \frac{(1-A_i)(Y_i - g(0,X_i))}{1-m(X_i)},
\]
and
\[\psi_a(W_i, \eta) := -1\]
which is also known as \abbreviation{AIPW}{augmented inverse propensity weighting} \citep{robins1995}.
Consequently, in this case, the nuisance \(\eta_0\) consists of \(\eta_0 = (g_0,\, m_0)\).

\subsection{Estimating Optimal Treatment Policies Based on Orthogonal Scores}
The methodology introduced in Section \ref{sec:DML} enables us to estimate average causal effects over all individuals.
In manufacturing settings, however, the magnitude of the effect of certain process steps strongly depends on the characteristics of each lot.
Given a set of the covariates $Z$, the CATE in the IRM model is defined as
$$\theta_0(z):=\mathbb{E}[g_0(1,X) - g_0(0,X) \,|\, Z=z].$$
\citet{semenova2021debiased} propose approximating $\theta_0(z)\approx b(z)^T\beta$ via a linear form, where $b(z)$ is a $d$-dimensional basis vector of $z$. 
The idea is based on projecting the part $\psi_b(W_i,\eta)$ of the Neyman-orthogonal scores onto the predefined basis vector $b(z)$. We can facilitate this estimation procedure to assign a counterfactual treatment policy as formulated in the problem set above (Section \ref{sec:problem}). To allow for a flexible conditional treatment effect, we use a B-spline basis for $b(z)$.
Because a positive CATE indicates improvements in yield, our estimated policy will be based on whether the effect on the yield surpasses some given threshold

$$\hat{\theta}(z) \ge c,$$

where $z$ is a defined subset of $x$, and the threshold $c>0$ can be used to incorporate costs and select rework groups at different effect levels.

An alternative approach is to directly estimate a policy based on the part $\psi_b(W_i,\eta)$ in Equation \ref{eq:psi}, as introduced by \citet{athey2021}.
The advantage of this method is a non-parametric CATE estimate, which can potentially incorporate more information of $X$ into $Z$.
The counterfactual treatment policy can be derived by solving
\begin{equation}
	\hat{\pi}=\argmax_{\pi\in\Pi}\frac{1}{n}\sum_{i=1}^{n}(2\pi(Z_i)-1)\psi_b(W_i,\hat{\eta}).
	\label{eq:policy}
\end{equation}
The learning problem in Equation \ref{eq:policy} can be reformulated as a weighted classification problem with weights $\lambda_i=|\psi_b(W_i,\hat{\eta})|$ and target $H_i = \textrm{sign}(\psi_b(W_i,\hat{\eta}))$. 
Given a specified policy class $\Pi$ and the corresponding regret
\begin{align*}
R(\pi):=\max_{\pi'\in\Pi}\left\{\mathbb{E}[Y_i(\pi'(Z_i))]\right\} - \mathbb{E}[Y_i(\pi(Z_i))],
\end{align*}
\cite{athey2021} are able to derive regret bounds of order $1/\sqrt{n}$.
To evaluate policies at different thresholds, we reduce the score by the desired threshold $\psi_b(W_i,\hat{\eta})-c$ before classification.

\subsection{Sensitivity of Estimation Towards Unobserved Confounding}\label{sec:longstoryshort}
In Section \ref{sec:causalinf}, we characterized each individual $i$ by its observable characteristics $X_i$.
However, this is only the ``short story''.
If Assumption \ref{ass:condex} fails due to an unobserved confounder $U$ that affects both treatment assignment and outcome, we might encounter a bias in the estimates.\\

\citet{chernozhukov2023long} provide a theory for estimating the effect of \textit{omitted variable bias} in the DML framework. Assume that in the observed data $W = (Y,A,X)$ the necessary confounding variable $U$ is not included in the features $X$. Because Assumption \ref{ass:condex} is violated, the procedure identifies a different parameter $\tilde{\theta}_0$ instead of the unconfounded $\theta_0$.
\citet{chernozhukov2023long} bound this bias as follows:
\begin{equation}
|\tilde{\theta}_0 -\theta_0|^2 \leq f(W, \, \rho, \, \zeta_y, \, \zeta_d)
\end{equation}\label{eq:sens}
where \(f\) is a known function and $\zeta_y$, $\zeta_d$, and $\rho$ are model-specific measures that capture the confounding strength of $U$.
In the setting of an IRM model, $\zeta_y$ denotes the nonparametric partial $R^2$ of $U$ with $Y$ given $(D,X)$.
$\zeta_d$ measures the relative gain in average conditional precision by including $U$ in the estimation of the treatment assignment.
Both measure the explanatory power of the unobserved confounder on either the outcome or treatment.
Lastly, $\rho$ refers to the correlation between the effects of the unobserved confounder $U$ on the outcome and the treatment mechanism.\\

In general, it is challenging to interpret and establish reasonable bounds for the strength of unobserved confounding $\zeta_y$, $\zeta_d$ and $\rho$.
However, to gauge the magnitude of the bounds, a popular approach is to rely on observed confounders to infer the strength of possibly unobserved confounders.
We can emulate omitted confounding by purposely omitting observed confounders from the adjustment set $X$ followed by re-estimating the whole model. This enables us to compare the ``long'' and ``short'' forms with and without omitted confounding.
From this comparison, one can estimate the explanatory power that the purposely omitted confounder had in the ``long'' and ``short'' forms, thereby gaining an idea of reasonable bounds for $\zeta_y$, $\zeta_d$ and $\rho$.
Of course, this procedure does not enable the estimation of the strength of the potential unobserved confounding $U$. \\

We then use Equation \ref{eq:sens} to assess the robustness of the estimation against a null hypothesis, $\Tilde{\theta}_0=0$.
The resulting robustness value ($\RV$) is defined as the required confounding strength ($\zeta_y=\RV$ and $\zeta_d=\RV$) such that the lower bound of the causal parameter includes the null hypothesis.
In other words, the robustness value represents the magnitude of an unobserved confounding effect necessary to reduce the estimated effect to zero.
In addition to the $\RV$ based on point estimates, robustness values can also be based on confidence intervals ($\RVa$), to assess which confounding is required to render the effect non-significant.

\section{Empirical Application}
In this section, we describe the application of our framework to the production process of
phosphor-converted white LEDs at AMS-Osram.

\subsection{Empirical Setup}\label{sec:data:setup}
\begin{figure}
\centering
  \scalebox{0.48}{\input{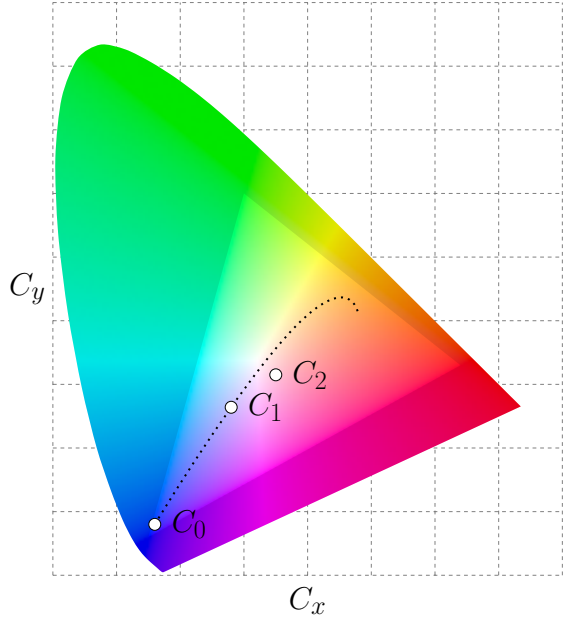}}
  \caption{
    CIE 1931 color space chromaticity diagram with exemplary color points
    \(C_0\) of the ground substrate,
    \(C_1\) after conversion, 
    and \(C_2\) for the final product.
  }
  \label{data:fig:ciexy}
\end{figure}

The manufacturing stages in LED production are commonly divided into frontend and backend processing.
The frontend involves the production of a monochromatic light-emitting semiconductor wafer,
whereas the backend involves the production steps after the separation of the individual chips.
After frontend processing, the different chips on the wafer are characterized
according to the emitted light spectra, sorted in conformity with the target product specification,
and glued onto a grid structure called a panel
\footnote{A panel holds \(28 \times 28\) chips and, for the purpose of our investigation, is regarded as a production lot.}
for further processing \((S_1)\).
After the sorting stage, bondwires are attached to the chips on the panel \((S_2)\),
enabling chipwise inline measurements.
These measurements take place during the phosphor conversion stage \((S_3)\),
during which the monochromatic blue light spectra of the chips
are shifted towards multichromatic white spectra.
Once conversion is complete, lenses are molded onto the chips \((S_4)\), changing the emitted light spectra again.
Finally, the chips are separated \((S_5)\) and subjected to a 100-percent inspection \((\FI)\),
which determines the share of conforming units on the panel \(Y\).
Conforming units are those for which the perceived color of the emitted spectrum matches the target color specification of the product.
The CIE 1931 chromaticity coordinate system (see \cite{Smith1931}) is used for this purpose,
mapping an observed spectrum to two-dimensional coordinates 
in the \(xy\)-chromaticity diagram (see Figure \ref{data:fig:ciexy}). 

Due to diverging production recipes further downstream,
it becomes necessary to restrict our investigation to a certain
product and thus to a certain sorting specification at stage \((S_1)\).
Based on our consultations with domain experts, we assume that influences from 
the backend are also mostly inhibited due to this restriction.

The rework decisions happen during the conversion stage \((S_3)\).
According to the recipe, a specific amount of phosphor slurry is sprayed onto the panel
followed by a curing step that finalizes the deposition of a single layer.
This procedure is repeated until the specified number of layers is reached, each with a defined target thickness.
Whereas the (fixed) phosphor mixture influences the \textit{conversion curve} (direction) in the color space,
the shift along this curve is mainly controlled by the layer thickness,
which can be adjusted through settings on the spray gun.
Figure \ref{data:fig:ciexy} depicts the shift from the unconverted monochromatic emission \(C_0\)
along a conversion curve towards the converted emission \(C_1\) 
after phosphor deposition according to the recipe, as well as the emission \(C_2\) after 100-percent inspection.
Subsequently, inline measurements at \(k=36\) different panel positions are taken,
revealing the color points \(C_1[j] = \big(C_{x}[j],\, C_{y}[j]\big)\) of the selected chips \(j=1, \ldots, k\).
The manufacturing stages \((S_1)\) and \((S_2)\) might introduce defectives,
resulting in invalid measurements \(C_1[i]\) as indicated by \(I[j] \in \{0,\,1\}\).
\footnote{
For example, contact issues after wirebonding \((S_2)\) are a reason for invalid measurements.
These issues persist until \((\FI)\), and because the rework decision at \((S_3)\) has no effect on it,
chips with invalid measurements are disregarded in the yield calculation.
Nevertheless, the number of defectives might still hold information about the unobserved lot-state \(P\).
}
Despite measuring individual chips, the operator is only shown the number of defectives \(I = \sum_{j=1}^k I[j]\)
and the mean panel color point
\[\overline{C}_1 = (\overline{C}_x,\, \overline{C}_y) = \frac{1}{k-I} \sum_{j=1}^k (1-I[j]) \big(C_x[j], \, C_y[j]\big)\]
upon which she decides whether to continue with production or to apply a further corrective layer.
The product state \(X_p\) observed by the operator consists solely of \(\overline{C}_1\) and \(I\).
The decision and outcome of corrective steps might be affected by the strain
the operator feels at the time of the decision-making.
As a proxy for this, we include the overall shopfloor workload \(V\) in the observed system state \(X_s\).
If an additional deposition step \((M)\) is issued, the inline inspection measurements \((\II)\)
and the decision-making \((\OR)\) are repeated with updated observed states as depicted in Figure \ref{model:fig:stage_flow}.
Based on the physical understanding of the process, it is assumed that knowledge of previous \(C_1\) measurements
yields no additional information, such that \(X_p\) consisting of the last measurement values is sufficient.

During times when the equipment is in an ``in-control'' state (\cite{Rosenblatt1986}),
no additional rework steps should be necessary to achieve a high yield.
But the sedimentation of phosphor particles, gradual curing of silicone components in the slurry,
and wear of the spray gun nozzle are examples of partially competing effects that lead to
``out-of-control'' states and subsequently to variations in the deposited phosphor.
Although applying an additional layer can counteract the depletion of phosphor particles
and shift the color point \(C_1\) further along the conversion curve towards the desired position,
depositing too many particles leads to overshooting, which cannot be corrected.

\subsection{Data} \label{Sec:Data}
The surveyed dataset includes \(N=69,046\) i.i.d. production lots of a specific product type manufactured at AMS-Osram.
It consists of the chip measurements before the rework decision,
their validity indicator, the shopfloor workload at the time of decision-making,
the decision itself, and the yield after production.
An overview of the variables can be found in Table \ref{tab:vars}.

\begin{table}[h]
  \centering
  \begin{tabular}{l|l}
    \toprule
    \textbf{Variable}                & \textbf{Description}  \\
    \midrule
    $C_{x}[1],\dots , C_{x}[36]$ & Individual chip $C_x$ measurements (lot probings)                         \\
    $C_{y}[1],\dots , C_{y}[36]$ & Individual chip $C_y$ measurements  (lot probings)                          \\
    $I[1],\dots , I[36]$ & Validity of \(C_x\) and \(C_y\) measurements                          \\
    $V$                  & Estimated workload (count of lots in the shift) \\
    $I$                  & Sum of invalid chip measurements                                 \\
    Yield $Y$            & $\%$ of chips usable at the end of the process $(Y \in [0,1])$ \\
    Rework decision $A$  & Indicator of treatment $(A \in \{0,1\})$         \\
    \bottomrule
  \end{tabular}
  \caption{Description of variables in the analysis}
  \label{tab:vars}
\end{table}

\subsection{Preprocessing}\label{sec:data:preprocessing}
We perform application-motivated preprocessing measures on the dataset.
The operator assigns the rework treatment by visually inspecting a plot of the covariates $\{(\overline{C}_{x}), (\overline{C}_{y})\}$ in the CIE chromaticity diagram.
However, as motivated in Section \ref{sec:data:setup}, the application of the conversion material
shifts the resulting color point along a conversion curve in the CIE color space (Figure \ref{data:fig:ciexy}).
The amount of conversion material applied controls the position on this curve,
while minor deviations orthogonal to it can be explained through process instabilities.
In the vicinity of the \(C_1\) measurements, the conversion curve can be well approximated by a linear function in the color space.
 Thus, the application of a \abbreviation{PCA}{principal component analysis} leads to transformed color coordinates 
 \(\PCA(C_1) = (C_m, \, C_s)\), where the primary principal component \(C_m\) captures the position on the conversion curve representing the main decision criterion,
and the secondary principal component \(C_s\) is mainly determined by process fluctuations.
In the following, we refer to \(C_m\) as the \textit{main color point measurement} and to \(C_s\) as the \textit{secondary color point measurement}.
The same notation applies to the mean
\(\PCA\left(\overline{C}_1\right) = \left(\overline{C}_m, \, \overline{C}_s\right) \)
and area mean color points
\(\PCA\left(\overline{C}_{1,l}\right) = \left(\overline{C}_{m,l}, \, \overline{C}_{s,l}\right)\).
For better explainability, we will work with these measures in the remainder of this paper.

Furthermore, we restrict our attention to a region on the conversion curve in which the treatment decision is not obvious.
This means that we subsample from the original dataset, rejecting data points with \(\overline{C}_m\) outside the interval $[\overline{C}_{m, min}, \overline{C}_{m, max}]$.
This interval is chosen in agreement with domain experts: observations exceeding \(\overline{C}_{m, max}\) 
should definitely be reworked, whereas data points lower than \(\overline{C}_{m, min}\) definitely should not.
While this procedure reduces the sample size to \(N=47,582\) observations,
it also helps to guarantee the overlap assumption (Assumption \ref{ass:olap}).
In the Appendix, we provide more details about the subsampling process (Figure \ref{fig:subsampling}). 
In the remainder of this paper, we restrict our attention to the reduced sample.

\subsection{Positivity Assumption}
\begin{figure}
\includegraphics[width=\linewidth]{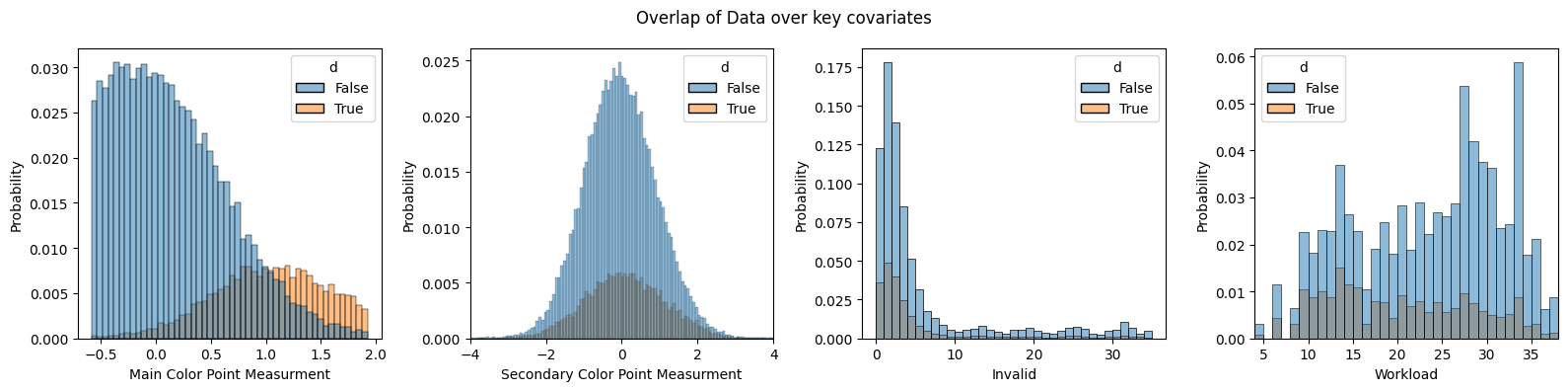}
\caption{Histogram of observations in the ``treatment'' and ``no treatment'' groups of different covariates. In general, the overlap assumption appears to hold.}\label{fig:overlap}
\end{figure}

In Figure \ref{fig:overlap}, we show that after subsampling there is a sufficient overlap of production lots with similar characteristics after conversion but with different treatments assigned.

To further assess the overlap, we follow the approach of \cite{McCaffrey2013}, which is also used by \cite{Ellickson2023}. \cite{McCaffrey2013} propose calculating the \abbreviation{PSB}{propensity score balance} given by
\begin{equation}
PSB_{A=a} = \frac{|\bar{X}^{(j)}_{A=a}-\bar{X}^{(j)}|}{\sigma^2_{X^{(j)}}}
\end{equation}
with 
\begin{equation}
  \bar{X}^{(j)}_{A=a} = \frac{\sum_{i=1}^n \mathbbm{1}(A_i=a)X_i^{(j)}/\hat{m}_0(X_i)}{\sum_{i=1}^n \mathbbm{1}(A_i=a)/\hat{m}_0(X_i)}.
\end{equation}
We report the scores for groups $A=0$ and $A=1$ in Appendix \ref{app:addresults}.
\cite{McCaffrey2013} consider balance scores lower than $0.2$ to be small,
a condition clearly met by all our balance scores.
Thus, we can confidently assert the overlap assumption.

\subsection{Average Treatment Effect and Comparison to Na\"ive Estimator}
We report the ATE and the ATT estimates in Table \ref{tab:ATE}.
Thereby we compare the naive ATE estimator, which uses the in-sample estimate of \(\mathbb{E}[Y\,|\,A=1] - \mathbb{E}[Y\,|\,A=0]\) with the causal estimate from an IRM. 
The naive ATE estimator indicates a significant negative effect amounting to a \(5.8\%\) of yield loss,
whereas the IRM estimator indicates a slight increase in yield of \(0.9\%\) over all production lots.
This difference is attributed to the unadjusted confounding in the naive estimation.
According to the naive estimation, reworking all lots would 
vastly reduce the share of items that meet specifications,
while the IRM estimator suggests only a slight increase in yield.
This slight increase is in line with our intuition that not every lot should undergo the rework step because,
some lots might be damaged depending on the product and system state.
This is further supported by the average treatment effect on the treated, which is significantly positive. 
In the group of reworked lots, we estimate an increase of $4.2\%$ in yield (indicating a decrease of the same amount if we had not reworked these lots).
These results underline the importance of a well-designed decision policy.
\begin{table}[]
  \centering
  \begin{tabular}{llrrrr}
  \toprule
       & estimator & coef.    & std. err. & 2.5\%      & 97.5\%     \\
       \midrule 
    ATE & naive  &  -0.058490 & 0.01005327  & -0.078195  & -0.038786 \\
    ATE & IRM & 	0.009241	&0.002753  &  0.003846  &	0.014636\\
    ATT & IRM & 	0.041827	&0.005952  &  0.030163  &	0.053492 \\
  \bottomrule
  \end{tabular}
  \caption{ATE and ATT of the inline rework step in the observed data}
  \label{tab:ATE}
\end{table}

\section{Results}\label{Sec:results}
In this section, we report the results of our analysis.
\footnote{Our analysis relies on the \texttt{DoubleML}-package of \cite{bach2022doubleml, bach2021doubleml}}
Because the quality of policy estimation improves with high-quality estimates of the unknown nuisance elements $\eta_0=(g_0,\, m_0)$ (see \eqref{eq:g_0} and \eqref{eq:m_0}), we rely on tuned machine learning algorithms.
\footnote{For hyperparameter tuning, we rely on the \textit{FLAML} AutoML package \citep{wang2021flaml} with an optimization time of one hour.}
We employ five-fold crossfitting and clip the propensity score estimate $\hat{m}$ at $0.025$ and $0.975$, respectively.
To ensure robustness, we split and use $70\%$ of the data to estimate the policies in Sections \ref{sec:rescate} and \ref{sec:respolicy}.
In Sections \ref{sec:resvalue} and \ref{sec:ressens}, we use the remaining $30\%$ of unseen data to evaluate the performance of the policies in the production chain.\\

\subsection{Estimation of Conditional Causal Effects}\label{sec:rescate}
\begin{figure}[h]
\includegraphics[width=\linewidth]{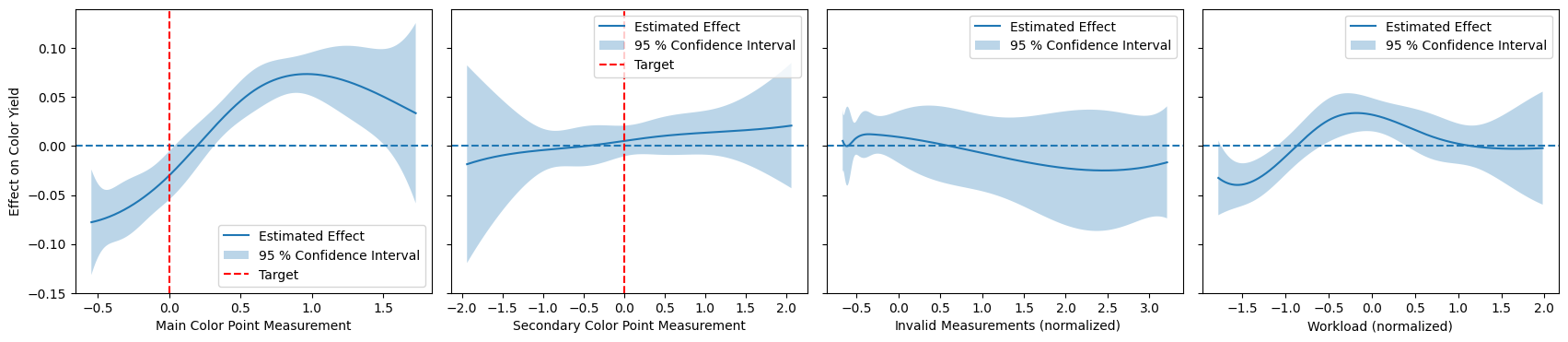}
\caption{Conditional average treatment effect estimated over different covariates which might be relevant for the rework policy. We observe conditional effects that significantly differ from the ATE only for the main measurement component and for lots that were produced under low workload.}\label{fig:cates}
\end{figure}

We project the orthogonal scores from the ATE estimation on different sets of covariates $Z$ to estimate the conditional effect.
As basis vectors $b(\tilde{z})$ for the CATE estimation, we construct cubic B-splines with five degrees of freedom.
Figure \ref{fig:cates} shows that the greatest effect variation can be observed for the mean main color point measurement \(\overline{C}_m\), which physically approximates the position on the color conversion curve.
In accordance with the physics of the process, there is a region in which chip lots (on average) have not reached the target color point and thus need additional layering.
In terms of the mean secondary color point measure \(\overline{C}_s\) and the count of invalid measurements per lot \(I\), we do not observe significant deviation of the conditional effects from the average effect.
For low workload \(V\), we observe a significant but small negative effect of the rework.
One reason might be that operators tend to experiment more under low workloads.
This means that panels that would not normally be subjected to rework under typical loads are reworked,
resulting in a negative effect.
Also, if the equipment is not used regularly, the setup of the machine before layer deposition might become more complicated 
(e.g., sedimentation of phosphor particles in the slurry reservoir of the machine),
affecting rework decisions and outcomes.
In the medium workload region, we see a peak in the color yield followed by a decline for higher workloads,
suggesting that while rework is reasonable to carry out,
it is more likely to be skipped due to time constraints.
We also estimate the conditional effect of two covariates simultaneously by using a tensor product of quadratic B-splines with five degrees of freedom in the two-dimensional case.
Due to the process physics, we opt to use the mean of the main and secondary color point measures \(\overline{C}_m\) and \(\overline{C}_s\) for the construction of the basis.
Figure \ref{fig:cate2d} shows that the sign of the conditional effect still largely depends on the main component. \\

\subsection{Optimal Treatment Policy Learning}\label{sec:respolicy}
\begin{figure}
    \centering
    \begin{minipage}{0.38\textwidth}
        \centering
        \includegraphics[width=\textwidth]{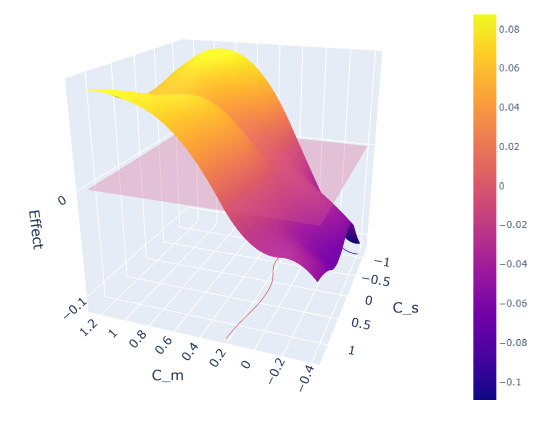} 
        \caption{Two-dimensional CATE evaluated on the mean main and secondary color point measure.}\label{fig:cate2d}
    \end{minipage}\hfill
    \begin{minipage}{0.58\textwidth}
        \centering
        \includegraphics[width=\textwidth]{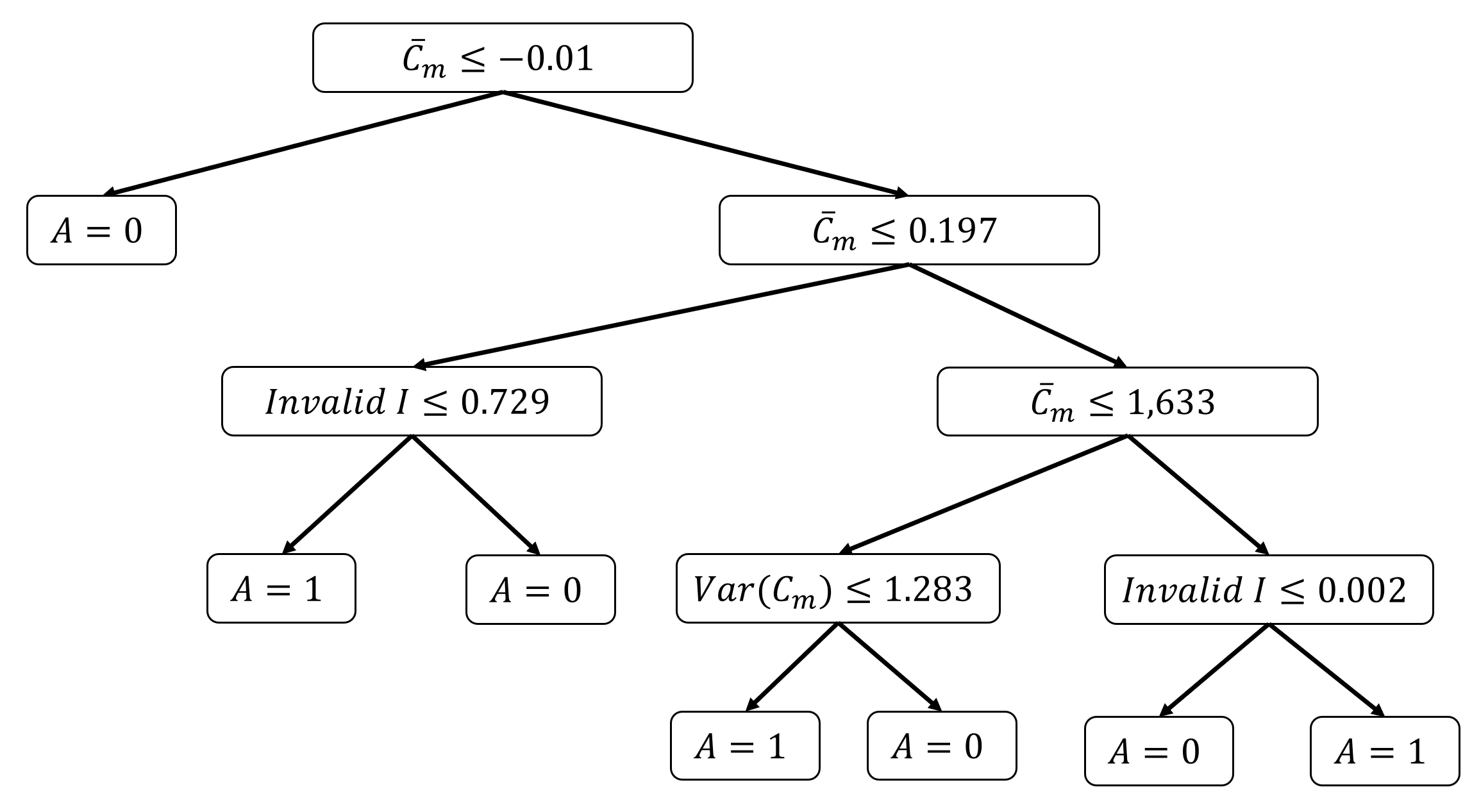} 
        \caption{Exemplary policy, optimized over primary and secondary color point measure statistics as well as the number of invalid measurements.}\label{fig:tree}
    \end{minipage}
\end{figure}

By assigning treatment in areas with a positive yield improvement for both the one-dimensional (depending solely on \(\overline{C}_m\)) and
the two-dimensional CATE (depending on \(\overline{C}_m\) and \(\overline{C}_s\)), we can derive simple policies based on only a few constraints.
To derive more complex policies, we employ the estimated score elements $\psi_b(W_i,\hat{\eta})$ and apply weighted classifiers as in Equation \ref{eq:policy} to obtain policy trees.
Figure \ref{fig:tree} shows an exemplary tree of depth four, in which the covariates $Z$ include the main and secondary color point measures as well as $I$ and the variance of $C_m[j]$ as an indicator for the individual distribution of quality in each lot.
Because an exact tree search at this depth is computationally expensive, we define the policy class to be
that of greedy classification trees.
However, for comparison, we perform an exact tree search at depth two, which results in a simpler policy. 

\subsection{Out-of-sample Assessment of Policy Value}\label{sec:resvalue}
We collect new, unseen data to evaluate the value of the policies derived above.\footnote{It is worth noting that these conclusions are valid only in our subsample.}
Table \ref{tab:values} displays the values for the different policies.
The results in Table \ref{tab:values} indicate a large improvement in value for all data-driven policies.
The observational policy has a value of $0.5\%$, suggesting that the assignment of the inline rework step
increases the overall production yield by half a percent.
The optimized policy is estimated to have a value of $2-3\%$, depending on the assumed costs $c$ and the methodology used.
\begin{table}[H]
  \centering
  \resizebox{\columnwidth}{!}{%
    \begin{tabular}{l|rrr|rrr|rrr}
      \toprule
       \multicolumn{1}{c}{}&\multicolumn{3}{c}{0\% costs}&\multicolumn{3}{c}{1\% costs}&\multicolumn{3}{c}{3\% costs}\\
       \midrule
        & 2.5 \%     & effect     & 97.5 \%  & 2.5 \%     & effect     & 97.5 \%& 2.5 \%     & effect     & 97.5 \% \\
        \midrule
      \textbf{Simple CATE}   &0.023734& 0.030639& 0.037544& 0.021211& 0.027866& 0.034522& 0.018011& 0.024025& 0.030040 \\
      \textbf{CATE 2D}	     &0.022258& 0.028881&	0.035505& 0.021881&	0.028478& 0.035074& 0.018818& 0.024876&	0.030933 \\
      \textbf{Greedy Tree}   &0.019566& 0.026718&	0.033870& 0.020436&	0.027434& 0.034433& 0.015950& 0.021449&	0.026947 \\
      \textbf{Exact Tree}    &0.013120& 0.019034&	0.024948& 0.016406&	0.021466& 0.026527& 0.012193& 0.015997&	0.019802 \\
      \textbf{observed value}&0.003334& 0.007232& 0.011129\\
      \bottomrule
    \end{tabular}
  }
  \caption{Values of the estimated policies at different assumed cost levels $c$.}
  \label{tab:values}
\end{table}

\begin{figure}
\centering
\includegraphics[width=\linewidth]{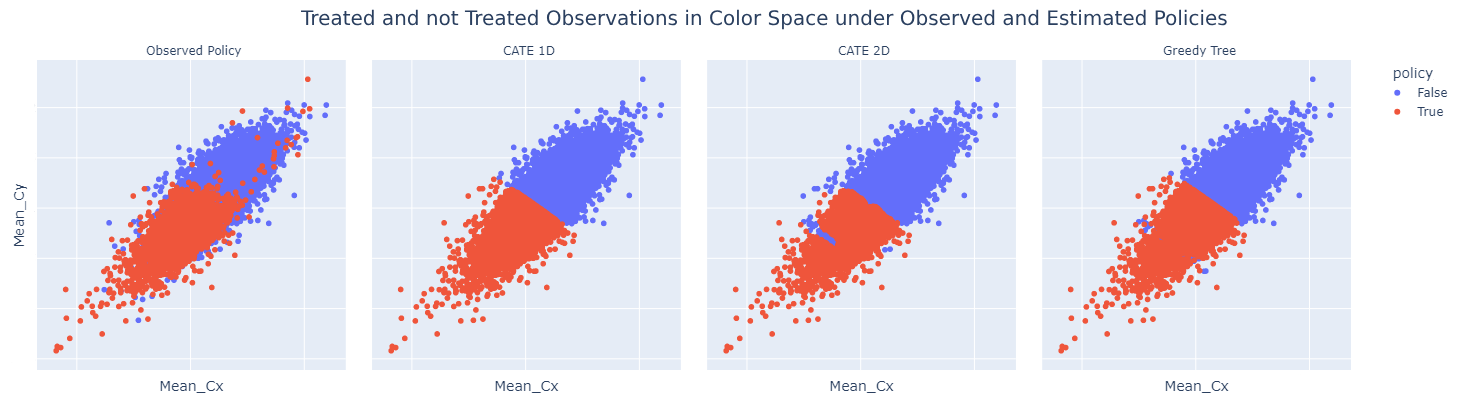}
\caption{Exemplary plot of the observed policy as well as the CATE 1D, CATE 2D and greedy policy tree policies (from left to right).}
\end{figure}

\subsection{Sensitivity and Robustness Considerations}\label{sec:ressens}
\begin{figure}
\centering
\includegraphics[width=.75\linewidth]{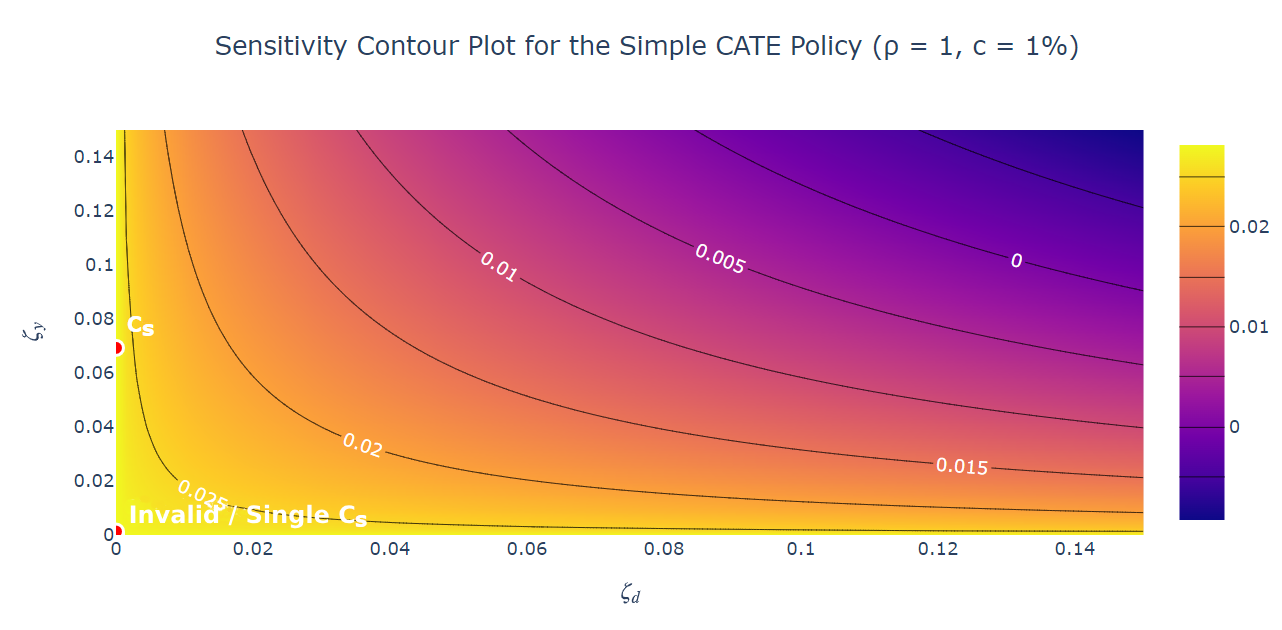}
\caption{Contour plot of different benchmarks. The labels showcase the estimated confounding strength explained by observed confounders in the sensitivity analysis.}\label{fig:sens}
\end{figure}

As motivated in Section \ref{sec:longstoryshort}, we aim to assess the possible influences of omitted variable bias.
Table \ref{tab:robvalues} reports the robustness values of the different policies against this bias.
To obtain comparable information to the robustness values,
we perform a benchmark on the one-dimensional CATE policy,
where we leave out different variables to estimate their impact if omitted.
In Table \ref{tab:sens}, it is evident that confounding similar to most of the variables we benchmarked would not substantially change the value estimation and is mostly smaller than the robustness value. The robustness values indicate that the estimated policy values are robust to (unobserved) confounding
effects similar in strength to the secondary color point measurement,
which we consider highly unlikely in our scenario.
Figure \ref{fig:sens} further emphasizes this point. 
\begin{table}[H]
\centering
\begin{tabular}{lrr}
\toprule
 & RV (\%) & RVa (\%)  \\
\midrule
Simple CATE       &11.836 &  9.161	 \\
CATE 2D         &11.726&  9.321 \\
Greedy Tree       &   9.107 &  7.095 \\
Exact Tree        & 16.945 &  11.258   \\
\bottomrule                      
\end{tabular}
\caption{Robustness values of the different policies against omitted variable bias.} \label{tab:robvalues}
\end{table}

\begin{table}[]
\centering
\begin{tabular}{l|rrrr|rrrr}
\toprule
\multicolumn{1}{c}{}& \multicolumn{4}{c}{\textbf{ATE}} & \multicolumn{4}{c}{\textbf{value}} \\
 & $\zeta_y$ & $\zeta_d$ & $\rho$    & $\Delta \theta_0$ & $\zeta_y$            & $\zeta_d$            & $\rho$               & $\Delta value$ \\
\midrule
\textbf{All $C_m[j]$}       & 0.1538  & 1.0000  & -0.0431 & -0.0099                             &                      &                      &                      &                                    \\
\textbf{One $C_m[j]$}       & 0.0009  & 0.0049  & 0.4567  & 0.0006                              &                      &                      &                      &                                    \\
\textbf{All $C_s[j]$}       & 0.0588  & 0.0713  & -0.1856 & -0.0077                             & 0.0693             & 0.0                  & -1.0000            & -0.0021                          \\
\textbf{One $C_s[j]$}       & 0.0003  & 0.0027  & 0.3303  & 0.0002                              & 0.0000             & 0.0                  & 1.0000              & 0.0000                           \\
\textbf{Workload \(V\)}             & 0.0388  & 0.1757  & 0.1329  & 0.0067                              & 0.0241             & 1.0                  & 0.125              & 0.0040                           \\
\textbf{Invalid \(I\)}              & 0.0000  & 0.0044  & 1.0000  & 0.0007                              & 0.0012             & 0.0                  & 1.00000              & 0.0014                          \\
\bottomrule
\end{tabular}
\caption{Sensitivity benchmark of the ATEs as well as values for the one-dimensional CATE policy with multiple covariates} \label{tab:sens}
\end{table}

\section{Conclusion}
In this paper, we have introduced a framework based on causal machine learning to derive optimal rework policies in lot-based manufacturing systems.
Whereas decision-making depends on the current product and system state,
we use the direct outcome of interest -- costs related to production yield -- as the optimization target.
We have also highlighted that confounding in our general rework setting can lead to severe bias when assessing rework decision-making.

Using a unique industry dataset, we have shown in our application that the causal machine learning approach
can significantly improve the value of the inline rework step in LED production.

The robustness and consistency of the different approaches support the argument that the main color measurements $C_m[j]$ contain most of the relevant information regarding a rework decision.
Thus, a policy based on this information should be implemented in production.
Our sensitivity analysis shows that this conclusion is robust against a reasonable amount of unobserved confounding.
Our results in Section \ref{sec:resvalue} consider only the subsample of approximately two-thirds of the data with sufficient overlap.
However, for the data points outside the considered interval, we expect no yield changes in the worst case.
Thus, we can conservatively estimate the global value to be at least 66
\% of that observed in our local estimates.\\

Although our approach addresses a wide range of questions, we would like to point out possible extensions.
Because production stages subsequent to the rework decision can influence the final yield without affecting the rework decision,
the causal setup described in Section \ref{model:sec:causal_setting} could be extended to include non-confounding system states stemming from these stages.

Such features -- for example, manufacturing parameters defined in the product recipe, or features that describe the condition of specific machines -- can help to explain yield variation.
Manufacturers tend to avoid fluctuations in production, such that recipe parameters are usually held constant after the product has been ramped up. Thus, the effect of recipe variation on yield, while certainly present, can hardly be estimated from production data alone. Furthermore, features that describe the machine condition have limited validity due to equipment wear or maintenance that occurs after the rework decision and before actual machine usage. In this regard, careful feature design is essential.

Our application simplifies the correction step to a binary decision; however, the framework could be adapted to allow for a set of specific or even parameterized repair decisions.
In our analysis, this would involve adjusting certain machine settings directly instead of deciding on fixed parameter settings for the rework step.
Furthermore, adjusting these settings directly during production at the cost of additional inspection steps between the layer deposition would more closely align with the ZDM approach.

In conclusion, it is important to combine insights from causal machine learning with the knowledge of domain experts to obtain value from the data. 
Having a clear and causally derived decision policy can help reduce the number of defective products and increase production yield.

\appendix
\section{Additional Results}\label{app:addresults}
\begin{figure}
\centering
\includegraphics[width=.6\linewidth]{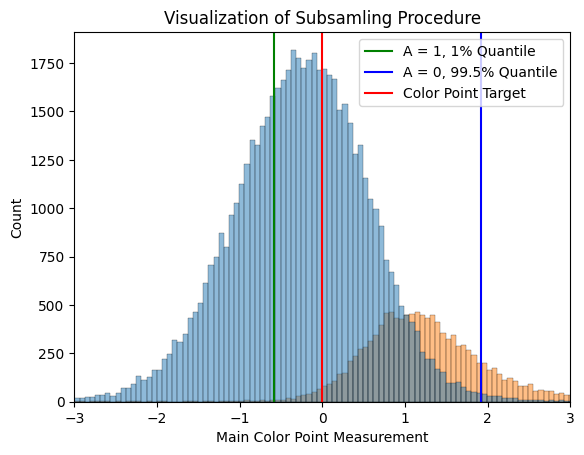}
\caption{Distributions of the main color point measurement by treatment and no-treatment group in the observed data. To ensure sufficient overlap, we clip the data for the analysis at the shown quantiles.}
\label{fig:subsampling}
\end{figure}

\FloatBarrier
In this section, we report additional tables and figures.\\ 
As motivated in Section \ref{sec:data:preprocessing}, we clip the data at the $1\%$ quantile of the treatment (rework) group and the $99.5\%$ quantile of the no-treatment group. Figure \ref{fig:subsampling} displays the cut-offs in the data.
Table \ref{tab:psb} reports the propensity score balance for all variables used as covariates in our analysis.
According to \cite{McCaffrey2013}, scores smaller than $0.2$ are considered to represent good overlap.
Thus, we can conclude that Assumption \ref{ass:olap} holds.
\begin{table}
  \centering
  \resizebox*{!}{\textheight}{%
    \begin{tabular}{ l r r }
      \toprule
      covariate& A=1	& A=0 \\
      \midrule
      $C_m[0]$ & 0.047534 & 0.038591  \\ 
      $C_m[1]$ & 0.076141 & 0.008915  \\ 
      $C_m[2]$ & 0.093222 & 0.010210  \\ 
      $C_m[3]$ & 0.051037 & 0.016896  \\ 
      $C_m[4]$ & 0.004794 & 0.040114  \\ 
      $C_m[5]$ & 0.080521 & 0.005825  \\ 
      $C_m[6]$ & 0.084074 & 0.005506  \\ 
      $C_m[7]$ & 0.053211 & 0.017469  \\ 
      $C_m[8]$ & 0.087347 & 0.003263  \\ 
      $C_m[9]$ & 0.090025 & 0.005484  \\ 
      $C_m[10]$ & 0.067338 & 0.013116  \\ 
      $C_m[11]$ & 0.096890 & 0.002121  \\ 
      $C_m[12]$ & 0.108802 & 0.002094  \\ 
      $C_m[13]$ & 0.065379 & 0.013240  \\ 
      $C_m[14]$ & 0.106175 & 0.003664  \\ 
      $C_m[15]$ & 0.099587 & 0.004157  \\ 
      $C_m[16]$ & 0.042630 & 0.023604  \\ 
      $C_m[17]$ & 0.082889 & 0.002739  \\ 
      $C_m[18]$ & 0.096174 & 0.004846  \\ 
      $C_m[19]$ & 0.059452 & 0.009946  \\ 
      $C_m[20]$ & 0.080126 & 0.002682  \\ 
      $C_m[21]$ & 0.090127 & 0.004553  \\ 
      $C_m[22]$ & 0.056841 & 0.013564  \\ 
      $C_m[23]$ & 0.070738 & 0.004782  \\ 
      $C_m[24]$ & 0.086241 & 0.008711  \\ 
      $C_m[25]$ & 0.067571 & 0.014752  \\ 
      $C_m[26]$ & 0.089712 & 0.005306  \\ 
      $C_m[27]$ & 0.096237 & 0.003064  \\ 
      $C_m[28]$ & 0.064272 & 0.016320  \\ 
      $C_m[29]$ & 0.058921 & 0.000978  \\ 
      $C_m[30]$ & 0.060912 & 0.010125  \\ 
      $C_m[31]$ & 0.007685 & 0.043738  \\ 
      $C_m[32]$ & 0.024386 & 0.016421  \\ 
      $C_m[33]$ & 0.044095 & 0.010250  \\ 
      $C_m[34]$ & 0.059248 & 0.012837  \\ 
      $C_m[35]$ & 0.008601 & 0.050252  \\ 
      $C_s[0]$ & 0.034725 & 0.003140  \\ 
      $C_s[1]$ & 0.036208 & 0.007336  \\ 
      $C_s[2]$ & 0.023773 & 0.003189  \\ 
      $C_s[3]$ & 0.050198 & 0.007174  \\ 
      $C_s[4]$ & 0.061728 & 0.002681  \\ 
      $C_s[5]$ & 0.034673 & 0.000273  \\ 
      $C_s[6]$ & 0.042859 & 0.001663  \\ 
      $C_s[7]$ & 0.066585 & 0.004676  \\ 
      $C_s[8]$ & 0.036237 & 0.004564  \\ 
      $C_s[9]$ & 0.052454 & 0.001202  \\ 
      $C_s[10]$ & 0.067758 & 0.006800  \\ 
      $C_s[11]$ & 0.050292 & 0.010142  \\ 
      $C_s[12]$ & 0.062176 & 0.001292  \\ 
      $C_s[13]$ & 0.054967 & 0.001960  \\ 
      $C_s[14]$ & 0.057509 & 0.004686  \\ 
      $C_s[15]$ & 0.064475 & 0.007918  \\ 
      $C_s[16]$ & 0.068183 & 0.017994  \\ 
      $C_s[17]$ & 0.068707 & 0.001435  \\ 
      $C_s[18]$ & 0.046259 & 0.007101  \\ 
      $C_s[19]$ & 0.040534 & 0.000799  \\ 
      $C_s[20]$ & 0.032138 & 0.014130  \\ 
      $C_s[21]$ & 0.060988 & 0.000473  \\ 
      $C_s[22]$ & 0.068694 & 0.001162  \\ 
      $C_s[23]$ & 0.064499 & 0.012974  \\ 
      $C_s[24]$ & 0.036732 & 0.006433  \\ 
      $C_s[25]$ & 0.058354 & 0.002368  \\ 
      $C_s[26]$ & 0.085918 & 0.008210  \\ 
      $C_s[27]$ & 0.091729 & 0.008323  \\ 
      $C_s[28]$ & 0.073288 & 0.003522  \\ 
      $C_s[29]$ & 0.083888 & 0.001078  \\ 
      $C_s[30]$ & 0.081795 & 0.010065  \\ 
      $C_s[31]$ & 0.044994 & 0.006911  \\ 
      $C_s[32]$ & 0.078706 & 0.013295  \\ 
      $C_s[33]$ & 0.057090 & 0.004444  \\ 
      $C_s[34]$ & 0.074954 & 0.010734  \\ 
      $C_s[35]$ & 0.053991 & 0.004596  \\ 
      Invalid $I$ & 0.066721 & 0.018125  \\ 
      Workload $V$ & 0.105923 & 0.017009 \\ 
      \bottomrule
    \end{tabular}%
  }
  \caption{Propensity Balance Scores for all covariates in our model}\label{tab:psb}
\end{table}
\FloatBarrier

\section*{Acknowledgements}
This work was funded by the Bavarian Joint Research Program (BayVFP) – Digitization (Funding reference: DIK0294/01). The research partners ams OSRAM and Economic AI kindly thank the VDI/VDE-IT Munich for the organization and the Free State of Bavaria for the financial support. We are grateful to Heribert Wankerl, Johannes Oberpriller, and Philipp Bach for their valuable comments. Additionally, we appreciate the interest and feedback from the participants of the 2023 KDD Workshop on Causal Discovery, Prediction, and Decision.
\singlespacing

\bibliographystyle{plainnat}
\bibliography{mybib}

\end{document}